\documentclass[sigconf]{acmart}

\usepackage{multirow}
\usepackage{url}
\AtBeginDocument{%
  \providecommand\BibTeX{{%
    \normalfont B\kern-0.5em{\scshape i\kern-0.25em b}\kern-0.8em\TeX}}}

\copyrightyear{2022}
\acmYear{2022}
\setcopyright{acmcopyright}\acmConference[MM '22]{Proceedings of the 30th ACM
International Conference on Multimedia}{October 10--14, 2022}{Lisboa, Portugal}
\acmBooktitle{Proceedings of the 30th ACM International Conference on Multimedia
(MM '22), October 10--14, 2022, Lisboa, Portugal}
\acmPrice{15.00}
\acmDOI{10.1145/3503161.3548329}
\acmISBN{978-1-4503-9203-7/22/10}

\begin{document}

\title{A Deep Learning based No-reference Quality Assessment Model for UGC Videos}


\author{Wei Sun}
\affiliation{%
  \institution{Shanghai Jiao Tong University}
  \city{Shanghai}
  \country{China}}
\email{sunguwei@sjtu.edu.cn}

\author{Xiongkuo Min}
\affiliation{%
  \institution{Shanghai Jiao Tong University}
  \city{Shanghai}
  \country{China}}
\email{minxiongkuo@sjtu.edu.cn}

\author{Wei Lu}
\affiliation{%
  \institution{Shanghai Jiao Tong University}
  \city{Shanghai}
  \country{China}}
\email{SJTU-Luwei@sjtu.edu.cn}

\author{Guangtao Zhai$^*$}
\affiliation{%
  \institution{Shanghai Jiao Tong University}
  \city{Shanghai}
  \country{China}}
\email{zhaiguangtao@sjtu.edu.cn}

\thanks{$^*$Corresponding author: Guangtao Zhai.}








\begin{abstract}
Quality assessment for User Generated Content (UGC) videos plays an important role in ensuring the viewing experience of end-users. Previous UGC video quality assessment (VQA) studies either use the image recognition model or the image quality assessment (IQA) models to extract frame-level features of UGC videos for quality regression, which are regarded as the sub-optimal solutions because of the domain shifts between these tasks and the UGC VQA task. In this paper, we propose a very simple but effective UGC VQA model, which tries to address this problem by training an end-to-end spatial feature extraction network to directly learn the quality-aware spatial feature representation from raw pixels of the video frames. We also extract the motion features to measure the temporal-related distortions that the spatial features cannot model. The proposed model utilizes very sparse frames to extract spatial features and dense frames (i.e. the video chunk) with a very low spatial resolution to extract motion features, which thereby has low computational complexity. With the better quality-aware features, we only use the simple multilayer perception layer (MLP) network to regress them into the chunk-level quality scores, and then the temporal average pooling strategy is adopted to obtain the video-level quality score. We further introduce a multi-scale quality fusion strategy to solve the problem of VQA across different spatial resolutions, where the multi-scale weights are obtained from the contrast sensitivity function of the human visual system. The experimental results show that the proposed model achieves the best performance on five popular UGC VQA databases, which demonstrates the effectiveness of the proposed model. The code is available at \url{https://github.com/sunwei925/SimpleVQA}.
\end{abstract}

\ccsdesc[500]{Computing methodologies~Modeling methodologies}

\keywords{video quality assessment, UGC videos, deep learning, feature fusion}


\maketitle

\section{Introduction}
\label{introduction}
With the proliferation of mobile devices and wireless networks in recent years, User Generated Content (UGC) videos have exploded over the Internet. It has become a popular daily activity for the general public to create, view, and share UGC videos through various social media applications such as YouTube, TikTok, etc. 
However, UGC videos are captured by a wide variety of consumers, ranging from professional photographers to amateur users, which makes the visual quality of UGC videos vary greatly. In order to ensure the Quality of Experience (QoE) of end-users, the service providers need to monitor the quality of UGC videos in the entire streaming media link, including but not limited to video uploading, compressing, post-processing, transmitting, etc. Therefore, with billions of video viewing and millions of newly uploaded UGC videos every day, an effective and efficient video quality assessment (VQA) model is needed to measure the perceptual quality of UGC videos.

Objective VQA can be divided into full-reference (FR), reduced-reference (RR), and no-reference (NR) according to the amount of pristine video information needed. Since there is no reference video for in-the-wild UGC videos, only NR VQA models are qualified for evaluating their quality. Although NR VQA algorithms \cite{saad2014blind,mittal2015completely,min2020study} have been studied for many years, most of them were developed for Professionally Generated Content (PGC) videos with synthetic distortions, where the pristine PGC videos are shot by photographers using professional devices and are normally of high quality, and the distorted PGC videos are then degraded by specific video processing algorithms such as video compression, transmission, etc. So, previous VQA studies mainly focus on modeling several types of distortions caused by specific algorithms, which makes them less effective for UGC videos with in-the-wild distortions. To be more specific, the emerging UGC videos pose the following challenges to the existing VQA algorithms for PGC videos:

First, the distortion types of UGC videos are diverse. A mass of UGC videos are captured by amateur users, which may suffer various distortion types such as under/over exposure, low visibility, jitter, noise, color shift, etc. These authentic distortions are introduced in the shooting processing and cannot be modeled by the single distortion type, which thereby requires that the VQA models have a more strong feature representation ability to qualify the authentic distortions.
Second, the content and forms of UGC videos are extremely rich. UGC videos can be natural scenes, animation \cite{wang2022subjective}, games \cite{zadtootaghaj2020quality, zadtootaghaj2018nr}, screen content, etc. Note that the statistics characteristics of different video content vary greatly. For example, the natural scenes statistics (NSS) features \cite{mittal2012making,mittal2012no,saad2014blind,mittal2015completely} are commonly used in the previous VQA studies to measure the distortions of natural scene content, but they may be ineffective for computer-generated content like animation or games. In addition, live videos, videoconferencing, etc. are also ubiquitous for UGC videos nowadays, whose quality is severely affected by the network bandwidth. 
Third, due to the advancement of shooting devices, more high resolution \cite{lu2022deep} and high frame rate \cite{madhusudana2021subjective, zheng2022no, zheng2022faver} videos have emerged on the Internet. The various kinds of resolutions and frame rates are also important factors for video quality. What's more, users can view the UGC videos through mobile devices anywhere and at any time, so the display \cite{rehman2015display} and the viewing environment such as ambient luminance \cite{sun2020dynamic}, etc. also affect the perceptual quality of UGC videos to a certain extent. However, these factors are rarely considered by previous studies.

The recently released large-scale UGC VQA databases such as KoNViD-1k \cite{hosu2017konstanz}, YouTube UGC \cite{wang2019youtube}, LSVQ \cite{ying2021patch}, etc. have greatly promoted the development of UGC VQA. Several deep learning based NR VQA models \cite{li2019quality,ying2021patch,wang2021rich,xu2021perceptual,li2021blindly} have been proposed to solve some challenges mentioned above and achieve pretty good performance. However, there are still some problems that need to be addressed. First, the previous studies either use the image recognition model \cite{li2019quality}\cite{ying2021patch} or the pretrained image quality assessment (IQA) models \cite{wang2021rich}\cite{xu2021perceptual}\cite{li2021blindly} to extract frame-level features, which lacks an end-to-end learning method to learn the quality-aware spatial feature representation from raw pixels of video frames. 
Second, previous studies usually extract the features from all video frames and have a very high computational complexity, making them difficult to apply to real-world scenarios. Since there is much redundancy spatial information between the adjacent frames, we argue that there is not necessary to extract the features from all frames. Third, the spatial resolution and frame rate of UGC videos as well as other factors such as the display, viewing environment, etc. are still rarely considered by these studies. However, these factors are very important for the perceptual quality of UGC videos since the contrast sensitivity of the human visual system (HVS) is affected by them.


\begin{figure*}[!t]
	\centering
	\includegraphics[height=2.2in]{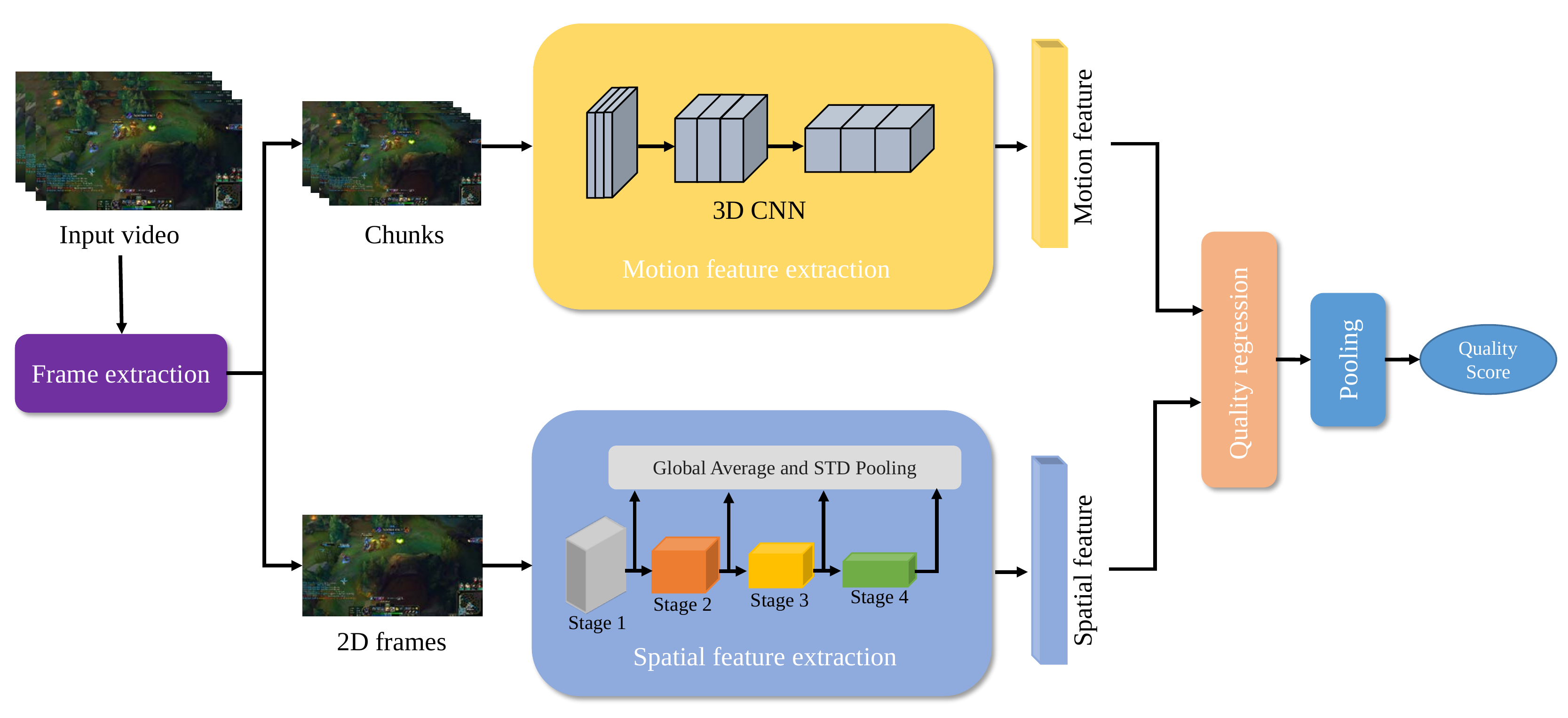}
	\caption{The network architecture of the proposed model. The proposed model contains the feature extraction module, the quality regression module, and the quality pooling module. The feature extraction module extracts two kinds of features, the spatial features and the motion features.}
	\label{model_framework}
\end{figure*}

In this paper, to address the challenges mentioned above, we propose a very simple but effective deep learning based VQA model for UGC videos. The proposed framework is illustrated in Figure \ref{model_framework}, which consists of the feature extraction module, the quality regression module, and the quality pooling module.
For the feature extraction module, we extract quality-aware features from the spatial domain and the spatial-temporal domain to respectively measure the spatial distortions and motion distortions. Instead of using the pretrained model to extract the spatial features in the previous studies, we propose to train an end-to-end spatial feature extraction network to learn quality-aware feature representation in the spatial domain, which thereby makes full use of various video content and distortion types in current UGC VQA databases.
We then utilize the action recognition network to extract the motion features, which can make up the temporal-related distortions that the spatial features cannot model. Considering that the spatial features are sensitive to the resolution while the motion features are sensitive to the frame rate, we first split the video into continuous chunks and then extract the spatial features and motion features by using a key frame of each chunk and all frames of each chunk but at a low spatial resolution respectively. So, the computational complexity of the proposed model can be greatly reduced.

For the quality regression module, we use the multilayer perception (MLP) network to map the quality-aware features into the chunk-level quality scores, and the temporal average pooling strategy is adopted to obtain the final video quality. In order to solve the problem of quality assessment across different resolutions, we introduce a multi-scale quality fusion strategy to fuse the quality scores of the videos with different resolutions, where the multi-scale weights are obtained from the contrast sensitivity function (CSF) of HVS by considering the viewing environment information. The proposed models are validated on five popular UGC VQA databases and the experimental results show that the proposed model outperforms other state-of-the-art VQA models by a large margin. What's more, the proposed model trained on a large-scale database such as LSVQ \cite{ying2021patch} achieves remarkable performance when tested on the other databases without any fine-tuning, which further demonstrates the effectiveness and generalizability of the proposed model.

In summary, this paper makes the following contributions:
\begin{enumerate}
	\item We propose an effective and efficient deep learning based model for UGC VQA, which includes the feature extraction module, the quality regression module, and the quality pooling module. The proposed model not only achieves remarkable performance on the five popular UGC VQA databases but also has a low computational complexity, which makes it very suitable for practical applications.
	\item The feature extraction module extracts two kinds of quality-aware features, the spatial features for spatial distortions and the spatial-temporal features for motion distortions, where the spatial features are learned from raw pixels of video frames via an end-to-end manner and the spatial-temporal features are extracted by a pretrained action recognition network. 
	\item We introduce a multi-scale quality fusion strategy to solve the problem of quality assessment across different resolutions, where the multi-scale weights are obtained from the contrast sensitivity function of the human visual system by considering the viewing environment information.
\end{enumerate}

\section{Related Work}
\label{related_work}
\subsection{Handcrafted feature based NR VQA Models}

A naive NR VQA method is to compute the quality of each frame via popular NR IQA methods such as NIQE \cite{mittal2012making}, BRISQUE \cite{mittal2012no}, CORNIA \cite{ye2012unsupervised} etc., and then pool them into the video quality score. A comparative study of various temporal pooling strategies on popular NR IQA methods can refer to \cite{tu2020comparative}. The temporal information is very important for VQA. V-BLIINDS \cite{saad2014blind} is a spatio-temporal natural scene statistics (NSS) model for videos by quantifying the NSS feature of frame-differences and motion coherency characteristics. 
Mittal \textit{et al.} \cite{mittal2015completely} propose a training-free blind VQA model named VIIDEO that exploits intrinsic statistics regularities of natural videos to quantify disturbances introduced due to distortions. 
TLVQM \cite{korhonen2019two} extracts abundant spatio-temporal features such as motion, jerkiness, blurriness, noise, blockiness, color, etc. at two levels of high and low complexity. VIDEVAL \cite{tu2021ugc} further combines the selected features from typical NR I/VQA methods to train a SVR model to regress them into the video quality. Since video content also affects its quality, especially for UGC videos, understanding the video content is beneficial to NR VQA. Previous handcrafted feature based methods are difficult to understand semantic information. Hence, some studies \cite{tu2021rapique, korhonen2020blind} try to combine the handcrafted features with the semantic-level features extracted by the pretrained CNN model to improve the performance of NR VQA models. For example, CNN-TLVQM \cite{korhonen2020blind} combines the handcrafted statistical temporal features from TLVQM and spatial features extracted by 2D-CNN model trained for IQA. RAPIQUE \cite{tu2021rapique} utilizes the quality-aware scene statistics features and semantics-aware deep CNN features to achieve a rapid and accurate VQA model for UGC videos.

\subsection{Deep learning based NR VQA Models}
With the release of several large-scale VQA databases \cite{hosu2017konstanz,wang2019youtube,ying2021patch}, deep learning based NR VQA models \cite{kim2018deep, li2019quality, ying2021patch, wang2021rich, xu2021perceptual, li2021blindly, sun2021deep, yi2021attention, cao2021deep} attract many researchers' attention. Liu \textit{et al.} \cite{liu2018end} propose a multi-task BVQA model V-MEON by jointly optimizing the 3D-CNN for quality assessment and compression distortion classification. VSFA \cite{li2019quality} first extracts the semantic features from a pre-trained CNN model and then uses a gated recurrent unit (GRU) network to model the temporal relationship between the semantic features of video frames. 
The authors of VSFA further propose MDVSFA \cite{li2021unified}, which trains the VSFA model on the multiple VQA databases to improve its performance and generalization. 
RIRNet \cite{chen2020rirnet} exploits the effect of motion information extracted from the multi-scale temporal frequencies for video quality assessment. 
Ying \textit{et al.} \cite{ying2021patch} propose a local-to-global region-based NR VQA model that combines the spatial features extracted from a 2D-CNN model and the spatial-temporal features from a 3D-CNN network. Wang \textit{et al.} \cite{wang2021rich} propose a feature-rich VQA model for UGC videos, which measures the quality from three aspects, compression level, video content, and distortion type and each aspect is evaluated by an individual neural network. 
Xu \textit{et al.} \cite{xu2021perceptual} first extract the spatial feature of the video frame from a pre-trained IQA model and use the graph convolution to extract and enhance these features, then extract motion information from the optical flow domain, and finally integrated the spatial feature and motion information via a bidirectional long short-term memory network. 
Li \textit{et al.} \cite{li2021blindly} also utilize the IQA model pre-trianed on multiple databases to extract quality-aware spatial features and the action recognition model to extract temporal features, and then a GRU network is used to model spatial and temporal features and regress them into the quality score.  Wen and Wang \cite{wen2021strong} propose a baseline I/VQA model for UGC videos, which calculates the video quality by averaging the scores of each frame and frame-level quality scores are obtained by a simple CNN network.

\section{Proposed Model}
\label{proposed_model}

The framework of the proposed NR VQA model is shown in Figure \ref{model_framework}, which consists of the feature extraction module, the quality regression module, and the quality pooling module. First, we extract the quality-aware features from the spatial domain and the spatial-temporal domain via the feature extraction module, which are utilized to evaluate the spatial distortions and motion distortions respectively. Then, the quality regression module is used to map the quality-aware features into chunk-level quality scores. Finally, we perform the quality pooling module to obtain the video quality score.

\subsection{Feature Extraction Module}
\label{feature_extraction_module}
In this section, we expect to extract the quality-aware features that can represent the impact of various distortion types and content on visual quality. The types of video distortion can be roughly divided into two categories: the spatial distortions and the motion distortions. The spatial distortions refer to the artifacts introduced in the video frames, such as noise, blur, compression, low visibility, etc. The motion distortions refer to the jitter, lagging due, etc., which are mainly caused by unstable shooting equipment, fast-moving objects, the low network bandwidth, etc. Therefore, we need to extract the quality-aware features from these two aspects.

Note that the characteristics of the spatial features and motion features are quite different. The spatial features are sensitive to the video resolution but insensitive to the video frame rate since the adjacent frames of the video contain lots of redundancy spatial information and higher resolution can represent more abundant high-frequency information, while motion features are the opposite because the motion distortions are reflected on the temporal dimension and these features are usually consistent for local regions of the frames.

Therefore, considering these characteristics, given a video $V$, whose number of frames and frame rate are $l$ and $r$ respectively, we first split the video $V$ into $N_c$ continuous chunks $c = \{c_i\}^{N_c}_{i=1}$ at an time interval $\tau$, where $N_c = l/(r*\tau)$, and there are $N_f = r*\tau$ frames in each chunk $c_i$, which is denoted as $c_i = \{x_{i,j}\}^{N_f}_{j=1}$. Then we only choose a key frame $x_{i,key}$ in each chunk to extract the spatial features and the motion features of each chunk are extracted using all frames in $c_i$ but at a very low spatial resolution. As a result, we can greatly reduce the computation complexity of the VQA model with little performance degradation.

\subsubsection{Spatial Feature Extraction Module}
\label{spatial_feature_extraction_module}


Given a frame $x$, we denote $f_w(x)$ as the output of the CNN model $f$ with trainable parameters $w=\{w_k\}$ applied on the frame $x$. Assume that there are $N_s$ stages in the CNN model, and $f_w^k(x)$ is the output feature maps extracted from the $k$-th stage, where $f_w^k(x)\in \mathbb{R}^{H_k \times W_k \times C_k}$, and $H_k$, $W_k$, and $C_k$ are the height, width, and the number of channels of the feature maps $f_w^k(x)$ respectively. In the following, we use the $f_w^k$ to replace the $f_w^k(x)$ for simplicity.

It is well known that the features extracted by the deep layers of the CNN model contain rich semantic information, and are suitable for representing content-aware features for UGC VQA. Moreover, previous studies indicate that the features extracted by the shallow layers of the CNN models contain low-level information \cite{zeiler2014visualizing, sun2019mc360iqa}, which responds to low-level features such as edges, corners, textures, etc. The low-level information is easily affected by the distortion and is therefore distortion-aware. Hence, we extract the quality-aware features via calculating the global mean and stand deviation of feature maps extracted from all stages of the CNN model.
Then, we apply global average and stand deviation pooling operations on the feature maps $f_{w}^{k}$:

\begin{equation}
\begin{aligned}
\mu_{f_{w}^{k}} &= { \rm GP_{avg}}(f_{w}^{k}), \\
\sigma_{f_{w}^{k}} &= { \rm GP_{std}}(f_{w}^{k}), \\
\end{aligned}
\end{equation}
where $\mu_{f_{w}^{k}}$ and $\sigma_{f_{w}^{k}}$ are the global means and stand deviation of feature maps $f_{w}^{k}$ respectively.
Finally, we concatenate the $\mu_{f_{w}^{k}}$ and $\sigma_{f_{w}^{k}}$ to derive the spital feature representation of our NR VQA model:
\begin{equation}
\begin{aligned}
F_{s}^k &= {\rm cat} ([\mu_{f_{w}^{k}}, \sigma_{f_{w}^{k}}]), \\
F_{s} &= {\rm cat} (\{F_{s}^k\}_{k=1}^{N_s}). \\
\end{aligned}
\end{equation}

\begin{table*}
\small
	\centering
	\renewcommand{\arraystretch}{1}
	\caption{Summary of the benchmark UGC VQA databases. Time duration: Seconds.}
	\label{the_database}
	\begin{tabular}{c|cccccccc}
		\hline
		\hline
		Database & Videos & Scenes & Resolution & Time Duration & Format & Distortion Type & DATA & Environment \\
		\hline
		KoNViD-1k \cite{hosu2017konstanz} & 1,200 & 1,200 & 540p & 8 & MP4 & Authentic & MOS + $\sigma$ & Crowd \\
		YouTube-UGC \cite{wang2019youtube} & 1500 & 1500 & 360p-4K & 20 & YUV, MP4 & Authentic & MOS + $\sigma$ & Crowd \\
		LSVQ \cite{ying2021patch} & 38,811 & 38,811 & 99p-4K & 5-12 & MP4 & Authentic & MOS + $\sigma$ & Crowd \\
		LBVD \cite{chen2019qoe} & 1,013 & 1,013 & 240p-540p & 10 & MP4 & Authentic, Transmission & MOS + $\sigma$ & In-lab \\
		LIVE-YT-Gaming \cite{yu2022subjective} & 600& 600& 360p-1080p& 8-9& MP4& Authentic& MOS & Crowd \\
		\hline
		\hline
	\end{tabular}
	
\end{table*}

\subsubsection{Motion Feature Extraction Module}
We extract the motion features as the complementary quality-aware features since UGC videos are commonly degraded by the motion distortions caused by the unstable shooting equipment or low bit rates in the living streaming or videoconferencing. The spatial features are difficult to handle these distortions because they are extracted by the intra-frames while motion distortions occur in the interframes. Therefore, the motion features are also necessary for evaluating the quality of UGC videos. Here, we utilize the pretrained action recognition model as the motion feature extractor to obtain the motion features of each video chunk. The action recognition model is designed to detect different kinds of action classes, so the feature representation of the action recognition network can reflect the motion information of the video to a certain extent. Therefore, given the video chunk $c$ and the action recognition network ${\rm MOTION}$, we can obtain the motion features:
\begin{equation}
\begin{aligned}
F_{m} = {\rm MOTION(c)}
\end{aligned}
\end{equation}
where $F_{m}$ represents the motion features extract by the action recognition network.

Therefore, given the video chunk $c$, we first select a key frame in the chunk to calculate the spatial features $F_{s}$. Then, we calculate the motion features $F_{m}$ using the whole frames but at a low spatial resolution in the video chunk. Finally, we obtain the quality-aware features for the video chunk $c$ by concatenating the spatial features and motion features:
\begin{equation}
\begin{aligned}
F = {\rm cat}([F_{s}, F_{m}]),
\end{aligned}
\end{equation}

\subsection{Quality Regression Module}
After extracting quality-aware feature representation by the feature extraction module, we need to map these features to the quality scores via a regression model. In this paper, we use the multi-layer perception (MLP) as the regression model to obtain the chunk-level quality due to its simplicity and effectiveness. The MLP consists of two fully connected layers and there are 128 and 1 neuron in each layer respectively. Therefore, we can obtain the chunk-level quality score via
\begin{equation}
\begin{array}{c}
q = f_{w_{\rm FC}}(F),
\end{array}
\end{equation}
where $ f_{w_{\rm FC}} $ denotes the function of the two FC layers and $q$ is the quality of the video chunk.


\subsection{Quality Pooling Module}
As stated in Section \ref{feature_extraction_module}, we split the video $V$ into $N_{c}$ continuous chunks $\{c_i\}_{i=1}^{N_c}$. For the chunk $c_i$, we can obtain its chunk-level quality score $q_i$ via the feature extraction module and the quality regression module. Then, it is necessary to pool the chunk-level scores into the video level. Though many temporal pooling methods have been proposed in literature \cite{tu2020comparative}\cite{li2019quality}, we find that the temporal averaging pooling achieves the best performance from Section \ref{quality_regression_module}. Therefore, the video-level quality is calculated as:
\begin{equation}
\begin{aligned}
Q = \frac{1}{N_c}\sum_{i=1}^{N_c} q_i,
\end{aligned}
\end{equation}
where $q_i$ is the quality of the $i$-th chunk and $Q$ is the video quality evaluated by the proposed model.

\subsection{Loss Function}
The loss function used to optimize the proposed models consists of two parts: the mean absolute error (MAE) loss and rank loss \cite{wen2021strong}. The MAE loss is used to make the evaluated quality scores close to the ground truth, which is defined as:
\begin{equation}
\begin{aligned}
L_{MAE} = \frac{1}{N}  \sum\limits_{i = 1}^{N} \left| Q_i- \hat{Q_i} \right|,
\end{aligned}
\end{equation}
where the $\hat{Q_i}$ is the ground truth quality score of the $i$-th video in a mini-batch and $N$ is the number of videos in the mini-batch.

The rank loss is further introduced to make the model distinguish the relative quality of videos better, which is very useful for the model to evaluate the videos with similar quality. Since the rank value between two video quality is non-differentiable, we use the following formula to approximate the rank value:
\begin{equation}
\begin{aligned}
L_{rank}^{ij} = \max(0, \left|\hat{Q_i} - \hat{Q_j}\right| - e(\hat{Q_i},\hat{Q_j})\cdot(Q_i - Q_j)),
\end{aligned}
\end{equation}
where $i$ and $j$ are two video indexes in a mini-batch, and $e(\hat{Q_i},\hat{Q_j})$ is formulated as:
\begin{equation}
\begin{array}{c}
e(\hat{Q_i},\hat{Q_j}) = \left\{
\begin{aligned}
1, \hat{Q_i} \geq \hat{Q_j}, \\
-1, \hat{Q_i} < \hat{Q_j}, \\
\end{aligned}
\right.
\end{array}
\end{equation}

Then, $L_{rank}$ is calculated via:
\begin{equation}
\begin{aligned}
L_{rank} = \frac{1}{N^2}\sum\limits_{i=1}^{N}\sum\limits_{j=1}^{N}L_{rank}^{ij}
\end{aligned}
\end{equation}

Finally, the loss function can be obtained by:
\begin{equation}
\begin{aligned}
L = L_{MAE} + \lambda \cdot L_{rank},
\end{aligned}
\end{equation}
where $\lambda$ is a hyper-parameter to balance the MAE loss and the rank loss.


\begin{table*}
\centering
\renewcommand{\arraystretch}{1}
\caption{Performance of the SOTA models and the proposed model on the KoNViD-1k, YouTube-UGC, LBVD, and LIVE-YT-Gaming databases. W.A. means the weight average results. The best performing model is highlighted in each column.}
\label{performance}
\begin{tabular}{c|c|cc|cc|cc|cc|cc}
\toprule[.15em]
\multirow{2}{*}{Type} & Database & \multicolumn{2}{c|}{KoNViD-1k} & \multicolumn{2}{c|}{YouTube-UGC} & \multicolumn{2}{c|}{LBVD} & \multicolumn{2}{c|}{LIVE-YT-Gaming} & \multicolumn{2}{c}{W.A.}\\
 & Criterion & SRCC & PLCC & SRCC & PLCC & SRCC & PLCC & SRCC & PLCC& SRCC & PLCC \\
\hline
\multirow{6}{*}{IQA}&NIQE & 0.542 & 0.553& 0.238& 0.278& 0.327& 0.387 & 0.280& 0.304& 0.359 & 0.393 \\
&BRISQUE & 0.657& 0.658& 0.382& 0.395& 0.435& 0.446& 0.604&0.638 & 0.513 &0.525 \\
&GM-LOG  & 0.658& 0.664& 0.368& 0.392& 0.314& 0.304& 0.312& 0.317& 0.433 &0.440 \\
&VGG19 & 0.774& 0.785& 0.703& 0.700& 0.676& 0.673& 0.678&0.658 & 0.714 &0.712 \\
&ResNet50 & 0.802& 0.810& 0.718& 0.710& 0.715& 0.717& 0.729& 0.768& 0.744 &0.751 \\
&KonCept512 & 0.735& 0.749& 0.587& 0.594& 0.626& 0.636& 0.643&0.649 & 0.650 &0.660 \\
\hline
\multirow{6}{*}{VQA}&V-BLIINDS  & 0.710& 0.704& 0.559& 0.555& 0.527& 0.558& 0.357& 0.403& 0.566 &0.578 \\
& TLVQM& 0.773& 0.769& 0.669& 0.659& 0.614& 0.590& 0.748& 0.756&  0.699 & 0.689 \\
&VIDEVAL & 0.783& 0.780& 0.779& 0.773& 0.707& 0.697& 0.807& 0.812& 0.766 &0.762 \\
&RAPIQUE & 0.803& 0.818& 0.759& 0.768& 0.712& 0.725& 0.803&0.825 & 0.767 &0.781 \\
&VSFA & 0.773& 0.775& 0.724& 0.743& 0.622& 0.642& 0.776&0.801& 0.721 &0.736  \\
&Li \textit{el al.} & 0.836& 0.834& 0.831& 0.819& -& -& -&- &- &-\\
& Pro. & \textbf{0.856} & \textbf{0.860}& \textbf{0.847} & \textbf{0.856} & \textbf{0.844}& \textbf{0.846}& \textbf{0.861} & \textbf{0.866} & \textbf{0.851}&\textbf{0.856} \\
\bottomrule[.15em]
\end{tabular}
\end{table*}

\subsection{Multi-scale Quality Fusion Strategy}
Previous studies evaluate the video quality either using the original spatial resolution or a fixed resized spatial resolution, which ignore that videos are naturally multi-scale \cite{zheng2022faver}. Some existing work \cite{wang2003multiscale}\cite{rehman2015display}\cite{min2017unified} shows that considering the multi-scale characteristics can improve the performance of image quality assessment. So, we propose a multi-scale quality fusion strategy to further improve the evaluation accuracy of the VQA model and this strategy is very useful to compare the quality of videos with different spatial resolutions.
\subsubsection{Multi-scale Video Quality Scores}
We first resize the resolution of the video into three fixed spatial scales, which are 540p, 720p, and 1080p, respectively. We do not downscale the video from the original scale to several lower resolution scales, which is a more common practice in previous studies. That is because when users watch videos in an application, the resolution of videos is actually adapted to the resolution of the playback device, and the modern display resolution is normally larger than 1080p. So, the perceptual quality of the low-resolution videos is also affected by the up-sampling artifacts, which also need to be considered by VQA models. Therefore, given a VQA model, we can derive three quality of videos at three scales, which are denoted as $Q_1$, $Q_2$, and $Q_3$ respectively. 
\subsubsection{Adaptive Multi-scale Weights}
The weight of each scale is obtained by considering the human psychological behaviors and the visual sensitivity characteristics. It is noted that the contrast perception ability of the HVS depends on the spatial frequency of the visual signal, which is modeled by the contrast sensitivity function (CSF). Specifically, we first define a viewing resolution factor $\xi$ as:
\begin{equation}
\begin{aligned}
\xi = \frac{\pi \cdot d \cdot n}{180 \cdot h_{s} \cdot 2},
\end{aligned}
\end{equation}
where the unit of $\xi$ is cycles per degree of visual angle (cpd), $d$ is the viewing distance (inch), $h_s$ is the height of the screen (inch), and $n$ denotes the number of pixels in the vertical direction of the screen. For the above three spatial scales of video, we can obtain the corresponding $\xi$, which are denoted as $\xi_1$, $\xi_2$, and $\xi_3$ respectively. We use $\xi$ to divide the spatial frequency range of the corresponding scale, which covers one section of the CSF formulated by:
\begin{equation}
\begin{aligned}
S(u) = \frac{5200e^{(-0.0016u^2(1+100/L)^{0.08})}}{\sqrt{(1+\frac{144}{X_0^2}+0.64u^2)(\frac{63}{L^{0.83}}+\frac{1}{1-e^(-0.02u^2)})}}
\end{aligned}
\end{equation}
where $u$, $L$, and $X_0^2$ indicate spatial frequency (cpd), luminance ($\rm {cd/m^2}$), and angular object area (squared degrees), respectively.

The weight of each scale is calculated as the area under the CSF within the corresponding frequency covering range:
\begin{equation}
\begin{aligned}
w_i = \frac{1}{Z} \int_{\xi_{i-1}}^{\xi_i} S(u){\rm d}u, i \in \{1,2,3\},
\end{aligned}
\end{equation}
where $i$ from 1 to 3 corresponds the finest to coarsest scale respectively, and $\xi_0$ corresponds the viewing resolution factor of 0. $Z$ is a normalization factor such that $\sum_i w_i = 1$.

Therefore, the multi-scale fusion quality score $Q_m$ is calculated as:
\begin{equation}
\begin{aligned}
Q_m = \prod_{i=1}^{3} Q_i^{w_i},
\end{aligned}
\end{equation}

\section{Experimental Validation}

\begin{table}
\centering
\renewcommand{\arraystretch}{1}
\caption{Performance of the SOTA models and the proposed models on the LSVQ database. Pro. M.S. refers to the proposed model implemented by the multi-scale quality fusion strategy. W.A. means the weighted average results. The best performing model is highlighted in each column.}
\label{performance_lsvq}
\begin{tabular}{c|cc|cc|cc}
\toprule[.15em]
 Database & \multicolumn{2}{c|}{Test} & \multicolumn{2}{c|}{Test-1080p} & \multicolumn{2}{c}{W.A.} \\
  Criterion & SRCC & PLCC & SRCC & PLCC& SRCC & PLCC  \\
\hline
TLVQM & 0.772&0.774& 0.589& 0.616& 0.712 & 0.722 \\
VIDEVAL & 0.794& 0.783& 0.545& 0.554& 0.712 &0.707 \\
VSFA & 0.801& 0.796& 0.675& 0.704 &0.759  &0.766 \\
PVQ & 0.827& 0.828& 0.711& 0.739 & 0.789 &0.799 \\
Li \textit{el al.} & 0.852& 0.854& \textbf{0.772}& 0.788 & 0.825 &0.832 \\
Pro. &  0.864& 0.861&  0.756&  0.801& 0.829 &0.841 \\
Pro. M.S. & \textbf{0.867}  &\textbf{0.861}  & 0.764  & \textbf{0.803}  & \textbf{0.833} & \textbf{0.842}  \\
\bottomrule[.15em]
\end{tabular}
\end{table}

\subsection{Experimental Protocol}
\subsubsection{Test Databases}
We test the proposed model on the five UGC VQA database: KoNViD-1k \cite{hosu2017konstanz}, YouTube-UGC \cite{wang2019youtube}, LSVQ \cite{ying2021patch},  LBVD \cite{chen2019qoe}, and LIVE-YT-Gaming \cite{yu2022subjective}. We summarize the main information of the databases in Table \ref{the_database}. The LSVQ database is the largest UGC VQA database so far, and there are 15 video categories such as animation, gaming, HDR, live music, sports, etc. in the YouTube-UGC database, which is more diverse than other databases. The LBVD database focuses on the live broadcasting videos, of which the videos are degraded by the authentic transmission distortions. The LIVE-YT-Gaming database consists of streamed gaming videos, where the video content is generated by computer graphics.

\subsubsection{Implementation Details}
We use the ResNet50 \cite{he2016deep} as the backbone of the spatial feature extraction module and the SlowFast R50 \cite{feichtenhofer2019slowfast} as the motion feature extraction model for the whole experiments. The weights of the ResNet50 are initialized by training on the ImageNet dataset \cite{deng2009imagenet}, the weights of the SlowFast R50 are fixed by training on the Kinetics 400 dataset \cite{kay2017kinetics}, and other weights are randomly initialized. For the spatial feature extraction module, we resize the resolution of the minimum dimension of key frames as 520 while maintaining their aspect ratios. In the training stage, the input frames are randomly cropped with the resolution of 448$\times$448. If we do not use the multi-scale quality fusion strategy, we crop the center patch with the same resolutions of 448$\times$448 in the testing stage. Note that we only validate the multi-scale quality fusion strategy on the model trained by the LSVQ database since there are enough videos with various spatial resolutions in it. For the motion feature extraction module, the resolution of the videos is resized to 224$\times$224 for both the training and testing stages. We use PyTorch to implement the proposed models. The Adam optimizer with the initial learning rate 0.00001 and batch size 8 are used for training the proposed model on a server with NVIDIA V100. The hyper-parameter $\lambda$ is set as 1. For simplicity, we select the first frame in each chunk as the key frame. For the multi-scale quality fusion strategy, there are $d = 35$, $n = 1080$, $h = 11.3$, $L = 200$, and $X^2_0 = 606$, and the final multi-scale weights for UGC videos are $w_1 = 0.8317$, $w_2 = 0.0939$, and $w_3 = 0.0745$.

\subsubsection{Comparing Algorithms}
We compare the proposed method with the following no-reference models:
\begin{itemize}
    \item IQA models: NIQE \cite{mittal2012making}, BRISQUE \cite{mittal2012no}, GM-LOG \cite{xue2014blind}, VGG19 \cite{simonyan2014very}, ResNet50 \cite{he2016deep}, and KonCept512 \cite{hosu2020koniq}.
    \item VQA models:  V-BLIINDS \cite{saad2014blind}, TLVQM \cite{korhonen2019two}, VIDEAL \cite{tu2021ugc}, RAPIQUE \cite{tu2021rapique}, VSFA \cite{li2019quality}, PVQ \cite{ying2021patch}, and Li \textit{et al.} \cite{li2021blindly}.
\end{itemize}

Since the number of videos in the LSVQ database is relatively large, we only compare some representative VQA models on the LSVQ database and omit the methods which perform poorly on the other four UGC databases.

\subsubsection{Evaluation Criteria}
We adopt two criteria to evaluate the performance of VQA models, which are Pearson linear correlation coefficient (PLCC) and Spearman rank-order correlation coefficient (SRCC).
PLCC reflects the prediction linearity of the VQA algorithm and SRCC indicates the prediction monotonicity. An excellent VQA model should obtain the value of SRCC and PLCC close to 1.
Before calculating the PLCC, we follow the same procedure in \cite{antkowiak2000final} to map the objective score to the subject score using a four-parameter logistic function.

For KoNViD-1k, YouTube-UGC, LBVD, and LIVE-YT-Gaming databases, we randomly split these databases into the training set with 80\% videos and the test set with 20\% videos for 10 times, and report the median values of SRCC and PLCC. For the LSVQ database, we follow the same training and test split suggested by \cite{ying2021patch} and report the performance on the test and test-1080p subsets.

\subsection{Performance Comparison with the SOTA Models}

The performance results of the VQA models on the KoNViD-1k, YouTube-UGC, LBVD, and LIVE-YT-Gaming databases are listed in Table \ref{performance}, and on the LSVQ database are listed in Table \ref{performance_lsvq}. From Table \ref{performance} and Table \ref{performance_lsvq}, we observe that the proposed model achieves the best performance on all five UGC VQA databases and leads by a large margin, which demonstrates that the proposed model does have a strong ability to measure the perceptual quality of various kinds of UGC videos. For the test-1080p subset of the LSVQ database, the proposed model is inferior to Li \textit{et al.}, which may be because the spatial resolution of most videos in the test-1080p subset is larger than 1080p while the proposed model resizes the spatial resolution of test videos into 448$\times$448, so the proposed model has a relatively poor ability to represent the characteristics of high-resolution videos. Through the multi-scale quality weighting fusion strategy, the proposed model can significantly improve the performance on the test-1080p subset. 

Then, most handcrafted feature based IQA models perform poorly on these UGC VQA databases especially for the LBVD and LIVE-YT-Gaming databases since they are designed for natural scene images with synthetic distortions and are difficult to handle the complex in-the-wild distortions and other video types such gaming, videoliving, etc. It is worth noting that through fine-tuning the deep CNN baseline i.e. ResNet50 on the VQA databases, it can achieve a pretty good performance, which also indicates that spatial features are very important for VQA tasks. For the NR VQA methods, the hand-crafted feature based NR VQA methods such as TLVQM and VIDEVAL achieve pretty well performance by incorporating the rich spatial and temporal quality features, such as NSS features, motion features, etc., but they are inferior to the deep learning based NR VQA methods due to the strong feature representation ability of CNN. VSFA extracts the spatial features from the pretrained image recognition model, which are not quality-aware, and achieves relatively poor performance when compared with other deep learning based methods. PVQ and Li \textit{et al.} methods both utilize the pretrained IQA model and ptretrained action recognition model to extract spatial and motion features respectively, and they perform better than other compared NR I/VQA methods but are inferior to the proposed model. 
Through training an end-to-end spatial feature extractor, the proposed model can take advantage of various video content and distortion types in the UGC databases and learn a better quality-aware feature representation. As a result, the proposed model achieves the best performance on all five UGC VQA databases.

\begin{table}
\centering
\renewcommand{\arraystretch}{1}
\caption{The results of ablation studies on the LSVQ database. S and M means the spatial features and motion features respectively, and S$^*$ means that the spatial features are extracted by the pretrained image classification network.}
\label{ablation_study}
\begin{tabular}{c|c|cc|cc}
\toprule[.15em]
\multirow{2}{*}{} & Database & \multicolumn{2}{c|}{Test} & \multicolumn{2}{c}{Test-1080p}  \\
 & Criterion & SRCC & PLCC & SRCC & PLCC  \\
\hline
\multirow{3}{*}{Feature}
&S$^*$+M& 0.847 & 0.841 & 0.732 &  0.774 \\
&S& 0.827 & 0.829 & 0.702 & 0.757  \\
&M& 0.660 & 0.669 & 0.569 & 0.621  \\
\hline
\multirow{2}{*}{Regression}
&GRU& 0.858 & 0.855 & 0.735 & 0.788  \\
&Transformer& 0.860 & 0.861 & 0.753 &0.799   \\
\hline
\multirow{2}{*}{Pooling}
& Method in \cite{li2019quality}& 0.860 & 0.858 & 0.733 & 0.786  \\
&1D CNN based& 0.864 & 0.862 & 0.739 &  0.790 \\
\bottomrule[.15em]
\end{tabular}
\end{table}

\subsection{Ablation Studies}
In this section, we conduct several ablation studies to investigate the effectiveness of each module in the proposed model, including the feature extraction module, and the quality regression module. All the experiments are tested on the LSVQ database since it is the largest UGC VQA model and is more representative.

\begin{table}
\small
\centering
\renewcommand{\arraystretch}{1}
\caption{The SRCC results of cross-database evaluation. The model is trained on the LSVQ database. }
\label{cross_database_evaluation}
\begin{tabular}{c|cccc}
\toprule[.15em]
 Database & KoNViD-1k & YouTube-UGC & LBVD & LIVE-YT-Gaming  \\
\hline
Pro.&  \textbf{0.860}&  0.789 &  0.689& 0.642  \\
Pro. M.S.&  0.859&  \textbf{0.822}&  \textbf{0.711}&  \textbf{0.683} \\
\bottomrule[.15em]
\end{tabular}
\end{table}

\begin{table*}
\centering
\renewcommand{\arraystretch}{1}
\caption{Comparison of computational complexity for the six VQA models and two proposed models. Time: Second.}
\label{computationsl_complexity}
\begin{tabular}{c|cccccccc}
\toprule[.15em]
 Methods & V-BLIINDS & TLVQM & VIDEVAL & VSFA& RAPIQUE & Li \textit{et al.} & Pro. & Pro. M.S. \\
\hline
Time & 61.982  & 219.992   & 561.408  &56.424  & 38.126  & 61.971 & \textbf{6.929}  & 8.448  \\
\bottomrule[.15em]
\end{tabular}
\end{table*}

\subsubsection{Feature Extraction Module}
The proposed model consists of the spatial feature extractor that learns the end-to-end spatial quality-aware features and the motion feature extractor that utilizes a pretrained action recognition model to represent motion information. 
Therefore, we first do not train the spatial feature extractor and directly use the weights trained on the ImageNet database to study the effect of the end-to-end training strategy for the spatial feature extractor. Then, we only use the end-to-end trained spatial features or the pretrained motion features to evaluate the quality of UGC videos to investigate the effect of these two kinds of features. The results are listed in Table \ref{ablation_study}. First, it is observed that the model using the motion features is inferior to the model using the spatial features and both of them are inferior to the proposed model, which indicates that both spatial and motion features are beneficial to the UGC VQA task and the spatial features are more important. Then, we find that end-to-end training for the spatial feature extractor can significantly improve the evaluation performance, which demonstrates that end-to-end trained spatial features represent better than that extracted by the pretrained image classification model.

\subsubsection{Quality Regression Module}
\label{quality_regression_module}
In this paper, we use the MLP as the regression model to derive the chunk-level quality scores. However, in previous studies, some sequential models such as GRU \cite{li2019quality}, Transformer \cite{li2021blindly}, etc. are also adopted to further consider the influence of the features extracted from adjacent frames. Here, we also adopt these methods as a comparison to investigate whether sequential models can improve the performance of the proposed models. Specifically, we replace the MLP module with the GRU and Transformer and keep other experimental setups the same. The results are listed in Table \ref{ablation_study}. We observe that models using GRU and Transformer are both inferior to the proposed model, which means that the MLP module is enough to regress the quality-aware features to quality scores though it is very simple. This conclusion is also consistent with \cite{wang2021rich}. The reason is that the proposed model and the model in \cite{wang2021rich} calculate the chunk-level quality score and the effect of adjacent frames are considered in the quality-aware features (i.e. motion features), while other VQA models \cite{li2019quality} \cite{li2021blindly} calculate the frame-level quality scores, which may need to consider the effect of adjacent frames in the quality regression module.


\subsubsection{Quality Pooling Module}
The proposed model uses the temporal average pooling method to fuse the chunk-level quality scores into the video level. It is noted that previous studies also propose several temporal pooling methods for VQA. 
In this section, we test two temporal pooling methods, which are the subjectively-inspired method introduced in \cite{li2019quality} and a learning based temporal pooling method using the 1D CNN. The results are listed in Table \ref{ablation_study}. From Table \ref{ablation_study}, we observe that the average pooling strategy achieves similar performance to the learning based pooling method, and both of them are superior to the subjectively-inspired methods. Since the average pooling strategy is simpler and does not increase the extra parameters, we use the temporal average pooling method in this paper.

\subsection{Cross-Database Evaluation}
UGC videos may contain various kinds of distortions and content, most of which may not exist in the training set. Hence, the generalization ability of the UGC VQA model is very important. In this section, we use the cross-database evaluation to test the generalization ability of the proposed model. Specifically, we train the proposed model on the LSVQ database and test the trained model on the other four UGC VQA databases. We list the results in Table \ref{cross_database_evaluation}. It is observed that the proposed model achieves excellent performance in cross-database evaluation. The SRCC results on the KoNViD-1k and YouTube-UGC databases both exceed 0.8, which have surpassed most VQA models trained on the corresponding database. We find that the multi-scale quality fusion strategy can significantly improve the performance on the databases containing videos with different spatial resolutions (YouTube-UGC, LBVD, and LIVE-YT-Gaming), which further demonstrates its effectiveness. It is also observed that the performance on the LBVD and LIVE-YT-Gaming databases is not good as the other two databases. The reason is that the LBVD and LIVE-YT-Gaming databases contain live broadcasting and gaming videos respectively, which may rarely exist in the LSVQ database. Since the single database can not cover all kinds of video types and distortions, we may further improve the generalization ability of the proposed model via the multiple database training strategy \cite{sun2021blind} \cite{zhang2021uncertainty} or the continual learning manner \cite{zhang2021continual} \cite{zhang2021task}.

\subsection{Computational Complexity}
The computational complexity is a very important factor that needs to be considered in practical applications. Hence, we test the computational complexity in this section. All models are tested on a computer with i7-6920HQ CPU, 16G RAM, and NVIDIA Quadro P400. The deep learning based models and the handcrafted based models are tested using the GPU and CPU respectively. We report the running time for a video with the resolution of 1920$\times$1080 and time duration of eight seconds in Table \ref{computationsl_complexity}. It is seen that the proposed model has a considerably low running time compared with other VQA models. The reason is that we use very sparse frames to calculate the spatial features while other deep learning based methods need dense frames. Moreover, we extract the motion features at a very low resolution, which only adds little computational complexity to the proposed model. The very low computational complexity makes the proposed model suitable for practical applications.

\section{Conclusion}
In this paper, we propose an effective and efficient NR VQA model for UGC videos. The proposed model extracts the quality-aware features from the spatial domain and the spatial-temporal domain to measure the spatial distortions and motion distortions respectively. We train the spatial feature extractor in an end-to-end training manner, so the proposed model can make full use of the various spatial distortions and content in the current VQA database. Then, the quality-aware features are regressed into the quality scores by the MLP network, and the temporal average pooling is used to obtain the video-level quality scores. We further introduce the multi-scale quality fusion strategy to address the problem of quality assessment across different spatial resolutions. The experimental results show that the proposed model can effectively measure the quality of UGC videos.

\begin{acks}
This work was supported by the National Natural Science Foundation of China (61831015, 61901260) and the National Key R\&D Program of China 2021YFE0206700.
\end{acks}

\bibliographystyle{ACM-Reference-Format}
\bibliography{sample-base}


\begin{thebibliography}{53}


\ifx \showCODEN    \undefined \def \showCODEN     #1{\unskip}     \fi
\ifx \showDOI      \undefined \def \showDOI       #1{#1}\fi
\ifx \showISBNx    \undefined \def \showISBNx     #1{\unskip}     \fi
\ifx \showISBNxiii \undefined \def \showISBNxiii  #1{\unskip}     \fi
\ifx \showISSN     \undefined \def \showISSN      #1{\unskip}     \fi
\ifx \showLCCN     \undefined \def \showLCCN      #1{\unskip}     \fi
\ifx \shownote     \undefined \def \shownote      #1{#1}          \fi
\ifx \showarticletitle \undefined \def \showarticletitle #1{#1}   \fi
\ifx \showURL      \undefined \def \showURL       {\relax}        \fi
\providecommand\bibfield[2]{#2}
\providecommand\bibinfo[2]{#2}
\providecommand\natexlab[1]{#1}
\providecommand\showeprint[2][]{arXiv:#2}

\bibitem[Antkowiak et~al\mbox{.}(2000)]%
        {antkowiak2000final}
\bibfield{author}{\bibinfo{person}{Jochen Antkowiak}, \bibinfo{person}{T~Jamal
  Baina}, \bibinfo{person}{France~Vittorio Baroncini}, \bibinfo{person}{Noel
  Chateau}, \bibinfo{person}{France FranceTelecom}, \bibinfo{person}{Antonio
  Claudio~Fran{\c{c}}a Pessoa}, \bibinfo{person}{F~Stephanie Colonnese},
  \bibinfo{person}{Italy~Laura Contin}, \bibinfo{person}{Jorge Caviedes}, {and}
  \bibinfo{person}{France Philips}.} \bibinfo{year}{2000}\natexlab{}.
\newblock \showarticletitle{Final report from the video quality experts group
  on the validation of objective models of video quality assessment march
  2000}.
\newblock  (\bibinfo{year}{2000}).
\newblock


\bibitem[Cao et~al\mbox{.}(2021)]%
        {cao2021deep}
\bibfield{author}{\bibinfo{person}{Yuqin Cao}, \bibinfo{person}{Xiongkuo Min},
  \bibinfo{person}{Wei Sun}, {and} \bibinfo{person}{Guangtao Zhai}.}
  \bibinfo{year}{2021}\natexlab{}.
\newblock \showarticletitle{Deep Neural Networks For Full-Reference And
  No-Reference Audio-Visual Quality Assessment}. In
  \bibinfo{booktitle}{\emph{2021 IEEE International Conference on Image
  Processing (ICIP)}}. IEEE, \bibinfo{pages}{1429--1433}.
\newblock


\bibitem[Chen et~al\mbox{.}(2019)]%
        {chen2019qoe}
\bibfield{author}{\bibinfo{person}{Pengfei Chen}, \bibinfo{person}{Leida Li},
  \bibinfo{person}{Yipo Huang}, \bibinfo{person}{Fengfeng Tan}, {and}
  \bibinfo{person}{Wenjun Chen}.} \bibinfo{year}{2019}\natexlab{}.
\newblock \showarticletitle{QoE evaluation for live broadcasting video}. In
  \bibinfo{booktitle}{\emph{2019 IEEE International Conference on Image
  Processing (ICIP)}}. IEEE, \bibinfo{pages}{454--458}.
\newblock


\bibitem[Chen et~al\mbox{.}(2020)]%
        {chen2020rirnet}
\bibfield{author}{\bibinfo{person}{Pengfei Chen}, \bibinfo{person}{Leida Li},
  \bibinfo{person}{Lei Ma}, \bibinfo{person}{Jinjian Wu}, {and}
  \bibinfo{person}{Guangming Shi}.} \bibinfo{year}{2020}\natexlab{}.
\newblock \showarticletitle{RIRNet: Recurrent-in-recurrent network for video
  quality assessment}. In \bibinfo{booktitle}{\emph{Proceedings of the 28th ACM
  International Conference on Multimedia}}. \bibinfo{pages}{834--842}.
\newblock


\bibitem[Deng et~al\mbox{.}(2009)]%
        {deng2009imagenet}
\bibfield{author}{\bibinfo{person}{Jia Deng}, \bibinfo{person}{Wei Dong},
  \bibinfo{person}{Richard Socher}, \bibinfo{person}{Li-Jia Li},
  \bibinfo{person}{Kai Li}, {and} \bibinfo{person}{Li Fei-Fei}.}
  \bibinfo{year}{2009}\natexlab{}.
\newblock \showarticletitle{Imagenet: A large-scale hierarchical image
  database}. In \bibinfo{booktitle}{\emph{2009 IEEE conference on computer
  vision and pattern recognition}}. Ieee, \bibinfo{pages}{248--255}.
\newblock


\bibitem[Feichtenhofer et~al\mbox{.}(2019)]%
        {feichtenhofer2019slowfast}
\bibfield{author}{\bibinfo{person}{Christoph Feichtenhofer},
  \bibinfo{person}{Haoqi Fan}, \bibinfo{person}{Jitendra Malik}, {and}
  \bibinfo{person}{Kaiming He}.} \bibinfo{year}{2019}\natexlab{}.
\newblock \showarticletitle{Slowfast networks for video recognition}. In
  \bibinfo{booktitle}{\emph{Proceedings of the IEEE/CVF international
  conference on computer vision}}. \bibinfo{pages}{6202--6211}.
\newblock


\bibitem[He et~al\mbox{.}(2016)]%
        {he2016deep}
\bibfield{author}{\bibinfo{person}{Kaiming He}, \bibinfo{person}{Xiangyu
  Zhang}, \bibinfo{person}{Shaoqing Ren}, {and} \bibinfo{person}{Jian Sun}.}
  \bibinfo{year}{2016}\natexlab{}.
\newblock \showarticletitle{Deep residual learning for image recognition}. In
  \bibinfo{booktitle}{\emph{Proceedings of the IEEE conference on computer
  vision and pattern recognition}}. \bibinfo{pages}{770--778}.
\newblock


\bibitem[Hosu et~al\mbox{.}(2017)]%
        {hosu2017konstanz}
\bibfield{author}{\bibinfo{person}{Vlad Hosu}, \bibinfo{person}{Franz Hahn},
  \bibinfo{person}{Mohsen Jenadeleh}, \bibinfo{person}{Hanhe Lin},
  \bibinfo{person}{Hui Men}, \bibinfo{person}{Tam{\'a}s Szir{\'a}nyi},
  \bibinfo{person}{Shujun Li}, {and} \bibinfo{person}{Dietmar Saupe}.}
  \bibinfo{year}{2017}\natexlab{}.
\newblock \showarticletitle{The Konstanz natural video database (KoNViD-1k)}.
  In \bibinfo{booktitle}{\emph{2017 Ninth international conference on quality
  of multimedia experience (QoMEX)}}. IEEE, \bibinfo{pages}{1--6}.
\newblock


\bibitem[Hosu et~al\mbox{.}(2020)]%
        {hosu2020koniq}
\bibfield{author}{\bibinfo{person}{Vlad Hosu}, \bibinfo{person}{Hanhe Lin},
  \bibinfo{person}{Tamas Sziranyi}, {and} \bibinfo{person}{Dietmar Saupe}.}
  \bibinfo{year}{2020}\natexlab{}.
\newblock \showarticletitle{KonIQ-10k: An ecologically valid database for deep
  learning of blind image quality assessment}.
\newblock \bibinfo{journal}{\emph{IEEE Transactions on Image Processing}}
  \bibinfo{volume}{29} (\bibinfo{year}{2020}), \bibinfo{pages}{4041--4056}.
\newblock


\bibitem[Kay et~al\mbox{.}(2017)]%
        {kay2017kinetics}
\bibfield{author}{\bibinfo{person}{Will Kay}, \bibinfo{person}{Joao Carreira},
  \bibinfo{person}{Karen Simonyan}, \bibinfo{person}{Brian Zhang},
  \bibinfo{person}{Chloe Hillier}, \bibinfo{person}{Sudheendra
  Vijayanarasimhan}, \bibinfo{person}{Fabio Viola}, \bibinfo{person}{Tim
  Green}, \bibinfo{person}{Trevor Back}, \bibinfo{person}{Paul Natsev},
  {et~al\mbox{.}}} \bibinfo{year}{2017}\natexlab{}.
\newblock \showarticletitle{The kinetics human action video dataset}.
\newblock \bibinfo{journal}{\emph{arXiv preprint arXiv:1705.06950}}
  (\bibinfo{year}{2017}).
\newblock


\bibitem[Kim et~al\mbox{.}(2018)]%
        {kim2018deep}
\bibfield{author}{\bibinfo{person}{Woojae Kim}, \bibinfo{person}{Jongyoo Kim},
  \bibinfo{person}{Sewoong Ahn}, \bibinfo{person}{Jinwoo Kim}, {and}
  \bibinfo{person}{Sanghoon Lee}.} \bibinfo{year}{2018}\natexlab{}.
\newblock \showarticletitle{Deep video quality assessor: From spatio-temporal
  visual sensitivity to a convolutional neural aggregation network}. In
  \bibinfo{booktitle}{\emph{Proceedings of the European Conference on Computer
  Vision (ECCV)}}. \bibinfo{pages}{219--234}.
\newblock


\bibitem[Korhonen(2019)]%
        {korhonen2019two}
\bibfield{author}{\bibinfo{person}{Jari Korhonen}.}
  \bibinfo{year}{2019}\natexlab{}.
\newblock \showarticletitle{Two-level approach for no-reference consumer video
  quality assessment}.
\newblock \bibinfo{journal}{\emph{IEEE Transactions on Image Processing}}
  \bibinfo{volume}{28}, \bibinfo{number}{12} (\bibinfo{year}{2019}),
  \bibinfo{pages}{5923--5938}.
\newblock


\bibitem[Korhonen et~al\mbox{.}(2020)]%
        {korhonen2020blind}
\bibfield{author}{\bibinfo{person}{Jari Korhonen}, \bibinfo{person}{Yicheng
  Su}, {and} \bibinfo{person}{Junyong You}.} \bibinfo{year}{2020}\natexlab{}.
\newblock \showarticletitle{Blind natural video quality prediction via
  statistical temporal features and deep spatial features}. In
  \bibinfo{booktitle}{\emph{Proceedings of the 28th ACM International
  Conference on Multimedia}}. \bibinfo{pages}{3311--3319}.
\newblock


\bibitem[Li et~al\mbox{.}(2021b)]%
        {li2021blindly}
\bibfield{author}{\bibinfo{person}{Bowen Li}, \bibinfo{person}{Weixia Zhang},
  \bibinfo{person}{Meng Tian}, \bibinfo{person}{Guangtao Zhai}, {and}
  \bibinfo{person}{Xianpei Wang}.} \bibinfo{year}{2021}\natexlab{b}.
\newblock \showarticletitle{Blindly Assess Quality of In-the-Wild Videos via
  Quality-aware Pre-training and Motion Perception}.
\newblock \bibinfo{journal}{\emph{arXiv preprint arXiv:2108.08505}}
  (\bibinfo{year}{2021}).
\newblock


\bibitem[Li et~al\mbox{.}(2019)]%
        {li2019quality}
\bibfield{author}{\bibinfo{person}{Dingquan Li}, \bibinfo{person}{Tingting
  Jiang}, {and} \bibinfo{person}{Ming Jiang}.} \bibinfo{year}{2019}\natexlab{}.
\newblock \showarticletitle{Quality assessment of in-the-wild videos}. In
  \bibinfo{booktitle}{\emph{Proceedings of the 27th ACM International
  Conference on Multimedia}}. \bibinfo{pages}{2351--2359}.
\newblock


\bibitem[Li et~al\mbox{.}(2021a)]%
        {li2021unified}
\bibfield{author}{\bibinfo{person}{Dingquan Li}, \bibinfo{person}{Tingting
  Jiang}, {and} \bibinfo{person}{Ming Jiang}.}
  \bibinfo{year}{2021}\natexlab{a}.
\newblock \showarticletitle{Unified quality assessment of in-the-wild videos
  with mixed datasets training}.
\newblock \bibinfo{journal}{\emph{International Journal of Computer Vision}}
  \bibinfo{volume}{129}, \bibinfo{number}{4} (\bibinfo{year}{2021}),
  \bibinfo{pages}{1238--1257}.
\newblock


\bibitem[Liu et~al\mbox{.}(2018)]%
        {liu2018end}
\bibfield{author}{\bibinfo{person}{Wentao Liu}, \bibinfo{person}{Zhengfang
  Duanmu}, {and} \bibinfo{person}{Zhou Wang}.} \bibinfo{year}{2018}\natexlab{}.
\newblock \showarticletitle{End-to-End Blind Quality Assessment of Compressed
  Videos Using Deep Neural Networks.}. In \bibinfo{booktitle}{\emph{ACM
  Multimedia}}. \bibinfo{pages}{546--554}.
\newblock


\bibitem[Lu et~al\mbox{.}(2022)]%
        {lu2022deep}
\bibfield{author}{\bibinfo{person}{Wei Lu}, \bibinfo{person}{Wei Sun},
  \bibinfo{person}{Xiongkuo Min}, \bibinfo{person}{Wenhan Zhu},
  \bibinfo{person}{Quan Zhou}, \bibinfo{person}{Jun He},
  \bibinfo{person}{Qiyuan Wang}, \bibinfo{person}{Zicheng Zhang},
  \bibinfo{person}{Tao Wang}, {and} \bibinfo{person}{Guangtao Zhai}.}
  \bibinfo{year}{2022}\natexlab{}.
\newblock \showarticletitle{Deep Neural Network for Blind Visual Quality
  Assessment of 4K Content}.
\newblock \bibinfo{journal}{\emph{arXiv preprint arXiv:2206.04363}}
  (\bibinfo{year}{2022}).
\newblock


\bibitem[Madhusudana et~al\mbox{.}(2021)]%
        {madhusudana2021subjective}
\bibfield{author}{\bibinfo{person}{Pavan~C Madhusudana},
  \bibinfo{person}{Xiangxu Yu}, \bibinfo{person}{Neil Birkbeck},
  \bibinfo{person}{Yilin Wang}, \bibinfo{person}{Balu Adsumilli}, {and}
  \bibinfo{person}{Alan~C Bovik}.} \bibinfo{year}{2021}\natexlab{}.
\newblock \showarticletitle{Subjective and objective quality assessment of high
  frame rate videos}.
\newblock \bibinfo{journal}{\emph{IEEE Access}}  \bibinfo{volume}{9}
  (\bibinfo{year}{2021}), \bibinfo{pages}{108069--108082}.
\newblock


\bibitem[Min et~al\mbox{.}(2017)]%
        {min2017unified}
\bibfield{author}{\bibinfo{person}{Xiongkuo Min}, \bibinfo{person}{Kede Ma},
  \bibinfo{person}{Ke Gu}, \bibinfo{person}{Guangtao Zhai},
  \bibinfo{person}{Zhou Wang}, {and} \bibinfo{person}{Weisi Lin}.}
  \bibinfo{year}{2017}\natexlab{}.
\newblock \showarticletitle{Unified blind quality assessment of compressed
  natural, graphic, and screen content images}.
\newblock \bibinfo{journal}{\emph{IEEE Transactions on Image Processing}}
  \bibinfo{volume}{26}, \bibinfo{number}{11} (\bibinfo{year}{2017}),
  \bibinfo{pages}{5462--5474}.
\newblock


\bibitem[Min et~al\mbox{.}(2020)]%
        {min2020study}
\bibfield{author}{\bibinfo{person}{Xiongkuo Min}, \bibinfo{person}{Guangtao
  Zhai}, \bibinfo{person}{Jiantao Zhou}, \bibinfo{person}{Mylene~CQ Farias},
  {and} \bibinfo{person}{Alan~Conrad Bovik}.} \bibinfo{year}{2020}\natexlab{}.
\newblock \showarticletitle{Study of subjective and objective quality
  assessment of audio-visual signals}.
\newblock \bibinfo{journal}{\emph{IEEE Transactions on Image Processing}}
  \bibinfo{volume}{29} (\bibinfo{year}{2020}), \bibinfo{pages}{6054--6068}.
\newblock


\bibitem[Mittal et~al\mbox{.}(2012a)]%
        {mittal2012no}
\bibfield{author}{\bibinfo{person}{Anish Mittal},
  \bibinfo{person}{Anush~Krishna Moorthy}, {and} \bibinfo{person}{Alan~Conrad
  Bovik}.} \bibinfo{year}{2012}\natexlab{a}.
\newblock \showarticletitle{No-reference image quality assessment in the
  spatial domain}.
\newblock \bibinfo{journal}{\emph{IEEE Transactions on image processing}}
  \bibinfo{volume}{21}, \bibinfo{number}{12} (\bibinfo{year}{2012}),
  \bibinfo{pages}{4695--4708}.
\newblock


\bibitem[Mittal et~al\mbox{.}(2015)]%
        {mittal2015completely}
\bibfield{author}{\bibinfo{person}{Anish Mittal}, \bibinfo{person}{Michele~A
  Saad}, {and} \bibinfo{person}{Alan~C Bovik}.}
  \bibinfo{year}{2015}\natexlab{}.
\newblock \showarticletitle{A completely blind video integrity oracle}.
\newblock \bibinfo{journal}{\emph{IEEE Transactions on Image Processing}}
  \bibinfo{volume}{25}, \bibinfo{number}{1} (\bibinfo{year}{2015}),
  \bibinfo{pages}{289--300}.
\newblock


\bibitem[Mittal et~al\mbox{.}(2012b)]%
        {mittal2012making}
\bibfield{author}{\bibinfo{person}{Anish Mittal}, \bibinfo{person}{Rajiv
  Soundararajan}, {and} \bibinfo{person}{Alan~C Bovik}.}
  \bibinfo{year}{2012}\natexlab{b}.
\newblock \showarticletitle{Making a “completely blind” image quality
  analyzer}.
\newblock \bibinfo{journal}{\emph{IEEE Signal processing letters}}
  \bibinfo{volume}{20}, \bibinfo{number}{3} (\bibinfo{year}{2012}),
  \bibinfo{pages}{209--212}.
\newblock


\bibitem[Rehman et~al\mbox{.}(2015)]%
        {rehman2015display}
\bibfield{author}{\bibinfo{person}{Abdul Rehman}, \bibinfo{person}{Kai Zeng},
  {and} \bibinfo{person}{Zhou Wang}.} \bibinfo{year}{2015}\natexlab{}.
\newblock \showarticletitle{Display device-adapted video quality-of-experience
  assessment}. In \bibinfo{booktitle}{\emph{Human Vision and Electronic Imaging
  XX}}, Vol.~\bibinfo{volume}{9394}. International Society for Optics and
  Photonics, \bibinfo{pages}{939406}.
\newblock


\bibitem[Saad et~al\mbox{.}(2014)]%
        {saad2014blind}
\bibfield{author}{\bibinfo{person}{Michele~A Saad}, \bibinfo{person}{Alan~C
  Bovik}, {and} \bibinfo{person}{Christophe Charrier}.}
  \bibinfo{year}{2014}\natexlab{}.
\newblock \showarticletitle{Blind prediction of natural video quality}.
\newblock \bibinfo{journal}{\emph{IEEE Transactions on Image Processing}}
  \bibinfo{volume}{23}, \bibinfo{number}{3} (\bibinfo{year}{2014}),
  \bibinfo{pages}{1352--1365}.
\newblock


\bibitem[Simonyan and Zisserman(2014)]%
        {simonyan2014very}
\bibfield{author}{\bibinfo{person}{Karen Simonyan} {and}
  \bibinfo{person}{Andrew Zisserman}.} \bibinfo{year}{2014}\natexlab{}.
\newblock \showarticletitle{Very deep convolutional networks for large-scale
  image recognition}.
\newblock \bibinfo{journal}{\emph{arXiv preprint arXiv:1409.1556}}
  (\bibinfo{year}{2014}).
\newblock


\bibitem[Sun et~al\mbox{.}(2019)]%
        {sun2019mc360iqa}
\bibfield{author}{\bibinfo{person}{Wei Sun}, \bibinfo{person}{Xiongkuo Min},
  \bibinfo{person}{Guangtao Zhai}, \bibinfo{person}{Ke Gu},
  \bibinfo{person}{Huiyu Duan}, {and} \bibinfo{person}{Siwei Ma}.}
  \bibinfo{year}{2019}\natexlab{}.
\newblock \showarticletitle{MC360IQA: a multi-channel CNN for blind 360-degree
  image quality assessment}.
\newblock \bibinfo{journal}{\emph{IEEE Journal of Selected Topics in Signal
  Processing}} \bibinfo{volume}{14}, \bibinfo{number}{1}
  (\bibinfo{year}{2019}), \bibinfo{pages}{64--77}.
\newblock


\bibitem[Sun et~al\mbox{.}(2020)]%
        {sun2020dynamic}
\bibfield{author}{\bibinfo{person}{Wei Sun}, \bibinfo{person}{Xiongkuo Min},
  \bibinfo{person}{Guangtao Zhai}, \bibinfo{person}{Ke Gu},
  \bibinfo{person}{Siwei Ma}, {and} \bibinfo{person}{Xiaokang Yang}.}
  \bibinfo{year}{2020}\natexlab{}.
\newblock \showarticletitle{Dynamic backlight scaling considering ambient
  luminance for mobile videos on lcd displays}.
\newblock \bibinfo{journal}{\emph{IEEE Transactions on Mobile Computing}}
  (\bibinfo{year}{2020}).
\newblock


\bibitem[Sun et~al\mbox{.}(2021a)]%
        {sun2021blind}
\bibfield{author}{\bibinfo{person}{Wei Sun}, \bibinfo{person}{Xiongkuo Min},
  \bibinfo{person}{Guangtao Zhai}, {and} \bibinfo{person}{Siwei Ma}.}
  \bibinfo{year}{2021}\natexlab{a}.
\newblock \showarticletitle{Blind quality assessment for in-the-wild images via
  hierarchical feature fusion and iterative mixed database training}.
\newblock \bibinfo{journal}{\emph{arXiv preprint arXiv:2105.14550}}
  (\bibinfo{year}{2021}).
\newblock


\bibitem[Sun et~al\mbox{.}(2021b)]%
        {sun2021deep}
\bibfield{author}{\bibinfo{person}{Wei Sun}, \bibinfo{person}{Tao Wang},
  \bibinfo{person}{Xiongkuo Min}, \bibinfo{person}{Fuwang Yi}, {and}
  \bibinfo{person}{Guangtao Zhai}.} \bibinfo{year}{2021}\natexlab{b}.
\newblock \showarticletitle{Deep learning based full-reference and no-reference
  quality assessment models for compressed ugc videos}. In
  \bibinfo{booktitle}{\emph{2021 IEEE International Conference on Multimedia \&
  Expo Workshops (ICMEW)}}. IEEE, \bibinfo{pages}{1--6}.
\newblock


\bibitem[Tu et~al\mbox{.}(2020)]%
        {tu2020comparative}
\bibfield{author}{\bibinfo{person}{Zhengzhong Tu}, \bibinfo{person}{Chia-Ju
  Chen}, \bibinfo{person}{Li-Heng Chen}, \bibinfo{person}{Neil Birkbeck},
  \bibinfo{person}{Balu Adsumilli}, {and} \bibinfo{person}{Alan~C Bovik}.}
  \bibinfo{year}{2020}\natexlab{}.
\newblock \showarticletitle{A comparative evaluation of temporal pooling
  methods for blind video quality assessment}. In
  \bibinfo{booktitle}{\emph{2020 IEEE International Conference on Image
  Processing (ICIP)}}. IEEE, \bibinfo{pages}{141--145}.
\newblock


\bibitem[Tu et~al\mbox{.}(2021a)]%
        {tu2021ugc}
\bibfield{author}{\bibinfo{person}{Zhengzhong Tu}, \bibinfo{person}{Yilin
  Wang}, \bibinfo{person}{Neil Birkbeck}, \bibinfo{person}{Balu Adsumilli},
  {and} \bibinfo{person}{Alan~C Bovik}.} \bibinfo{year}{2021}\natexlab{a}.
\newblock \showarticletitle{UGC-VQA: Benchmarking blind video quality
  assessment for user generated content}.
\newblock \bibinfo{journal}{\emph{IEEE Transactions on Image Processing}}
  (\bibinfo{year}{2021}).
\newblock


\bibitem[Tu et~al\mbox{.}(2021b)]%
        {tu2021rapique}
\bibfield{author}{\bibinfo{person}{Zhengzhong Tu}, \bibinfo{person}{Xiangxu
  Yu}, \bibinfo{person}{Yilin Wang}, \bibinfo{person}{Neil Birkbeck},
  \bibinfo{person}{Balu Adsumilli}, {and} \bibinfo{person}{Alan~C Bovik}.}
  \bibinfo{year}{2021}\natexlab{b}.
\newblock \showarticletitle{Rapique: Rapid and accurate video quality
  prediction of user generated content}.
\newblock \bibinfo{journal}{\emph{arXiv preprint arXiv:2101.10955}}
  (\bibinfo{year}{2021}).
\newblock


\bibitem[Wang et~al\mbox{.}(2022)]%
        {wang2022subjective}
\bibfield{author}{\bibinfo{person}{Tao Wang}, \bibinfo{person}{Zicheng Zhang},
  \bibinfo{person}{Wei Sun}, \bibinfo{person}{Xiongkuo Min},
  \bibinfo{person}{Wei Lu}, {and} \bibinfo{person}{Guangtao Zhai}.}
  \bibinfo{year}{2022}\natexlab{}.
\newblock \showarticletitle{Subjective Quality Assessment for Images Generated
  by Computer Graphics}.
\newblock \bibinfo{journal}{\emph{arXiv preprint arXiv:2206.05008}}
  (\bibinfo{year}{2022}).
\newblock


\bibitem[Wang et~al\mbox{.}(2019)]%
        {wang2019youtube}
\bibfield{author}{\bibinfo{person}{Yilin Wang}, \bibinfo{person}{Sasi Inguva},
  {and} \bibinfo{person}{Balu Adsumilli}.} \bibinfo{year}{2019}\natexlab{}.
\newblock \showarticletitle{YouTube UGC dataset for video compression
  research}. In \bibinfo{booktitle}{\emph{2019 IEEE 21st International Workshop
  on Multimedia Signal Processing (MMSP)}}. IEEE, \bibinfo{pages}{1--5}.
\newblock


\bibitem[Wang et~al\mbox{.}(2021)]%
        {wang2021rich}
\bibfield{author}{\bibinfo{person}{Yilin Wang}, \bibinfo{person}{Junjie Ke},
  \bibinfo{person}{Hossein Talebi}, \bibinfo{person}{Joong~Gon Yim},
  \bibinfo{person}{Neil Birkbeck}, \bibinfo{person}{Balu Adsumilli},
  \bibinfo{person}{Peyman Milanfar}, {and} \bibinfo{person}{Feng Yang}.}
  \bibinfo{year}{2021}\natexlab{}.
\newblock \showarticletitle{Rich features for perceptual quality assessment of
  UGC videos}. In \bibinfo{booktitle}{\emph{Proceedings of the IEEE/CVF
  Conference on Computer Vision and Pattern Recognition}}.
  \bibinfo{pages}{13435--13444}.
\newblock


\bibitem[Wang et~al\mbox{.}(2003)]%
        {wang2003multiscale}
\bibfield{author}{\bibinfo{person}{Zhou Wang}, \bibinfo{person}{Eero~P
  Simoncelli}, {and} \bibinfo{person}{Alan~C Bovik}.}
  \bibinfo{year}{2003}\natexlab{}.
\newblock \showarticletitle{Multiscale structural similarity for image quality
  assessment}. In \bibinfo{booktitle}{\emph{The Thrity-Seventh Asilomar
  Conference on Signals, Systems \& Computers, 2003}},
  Vol.~\bibinfo{volume}{2}. Ieee, \bibinfo{pages}{1398--1402}.
\newblock


\bibitem[Wen and Wang(2021)]%
        {wen2021strong}
\bibfield{author}{\bibinfo{person}{Shaoguo Wen} {and} \bibinfo{person}{Junle
  Wang}.} \bibinfo{year}{2021}\natexlab{}.
\newblock \showarticletitle{A strong baseline for image and video quality
  assessment}.
\newblock \bibinfo{journal}{\emph{arXiv preprint arXiv:2111.07104}}
  (\bibinfo{year}{2021}).
\newblock


\bibitem[Xu et~al\mbox{.}(2021)]%
        {xu2021perceptual}
\bibfield{author}{\bibinfo{person}{Jiahua Xu}, \bibinfo{person}{Jing Li},
  \bibinfo{person}{Xingguang Zhou}, \bibinfo{person}{Wei Zhou},
  \bibinfo{person}{Baichao Wang}, {and} \bibinfo{person}{Zhibo Chen}.}
  \bibinfo{year}{2021}\natexlab{}.
\newblock \showarticletitle{Perceptual Quality Assessment of Internet Videos}.
  In \bibinfo{booktitle}{\emph{Proceedings of the 29th ACM International
  Conference on Multimedia}}. \bibinfo{pages}{1248--1257}.
\newblock


\bibitem[Xue et~al\mbox{.}(2014)]%
        {xue2014blind}
\bibfield{author}{\bibinfo{person}{Wufeng Xue}, \bibinfo{person}{Xuanqin Mou},
  \bibinfo{person}{Lei Zhang}, \bibinfo{person}{Alan~C Bovik}, {and}
  \bibinfo{person}{Xiangchu Feng}.} \bibinfo{year}{2014}\natexlab{}.
\newblock \showarticletitle{Blind image quality assessment using joint
  statistics of gradient magnitude and Laplacian features}.
\newblock \bibinfo{journal}{\emph{IEEE Transactions on Image Processing}}
  \bibinfo{volume}{23}, \bibinfo{number}{11} (\bibinfo{year}{2014}),
  \bibinfo{pages}{4850--4862}.
\newblock


\bibitem[Ye et~al\mbox{.}(2012)]%
        {ye2012unsupervised}
\bibfield{author}{\bibinfo{person}{Peng Ye}, \bibinfo{person}{Jayant Kumar},
  \bibinfo{person}{Le Kang}, {and} \bibinfo{person}{David Doermann}.}
  \bibinfo{year}{2012}\natexlab{}.
\newblock \showarticletitle{Unsupervised feature learning framework for
  no-reference image quality assessment}. In \bibinfo{booktitle}{\emph{2012
  IEEE conference on computer vision and pattern recognition}}. IEEE,
  \bibinfo{pages}{1098--1105}.
\newblock


\bibitem[Yi et~al\mbox{.}(2021)]%
        {yi2021attention}
\bibfield{author}{\bibinfo{person}{Fuwang Yi}, \bibinfo{person}{Mianyi Chen},
  \bibinfo{person}{Wei Sun}, \bibinfo{person}{Xiongkuo Min},
  \bibinfo{person}{Yuan Tian}, {and} \bibinfo{person}{Guangtao Zhai}.}
  \bibinfo{year}{2021}\natexlab{}.
\newblock \showarticletitle{Attention Based Network For No-Reference UGC Video
  Quality Assessment}. In \bibinfo{booktitle}{\emph{2021 IEEE International
  Conference on Image Processing (ICIP)}}. IEEE, \bibinfo{pages}{1414--1418}.
\newblock


\bibitem[Ying et~al\mbox{.}(2021)]%
        {ying2021patch}
\bibfield{author}{\bibinfo{person}{Zhenqiang Ying}, \bibinfo{person}{Maniratnam
  Mandal}, \bibinfo{person}{Deepti Ghadiyaram}, {and} \bibinfo{person}{Alan
  Bovik}.} \bibinfo{year}{2021}\natexlab{}.
\newblock \showarticletitle{Patch-VQ:'Patching Up'the Video Quality Problem}.
  In \bibinfo{booktitle}{\emph{Proceedings of the IEEE/CVF Conference on
  Computer Vision and Pattern Recognition}}. \bibinfo{pages}{14019--14029}.
\newblock


\bibitem[Yu et~al\mbox{.}(2022)]%
        {yu2022subjective}
\bibfield{author}{\bibinfo{person}{Xiangxu Yu}, \bibinfo{person}{Zhenqiang
  Ying}, \bibinfo{person}{Neil Birkbeck}, \bibinfo{person}{Yilin Wang},
  \bibinfo{person}{Balu Adsumilli}, {and} \bibinfo{person}{Alan~C Bovik}.}
  \bibinfo{year}{2022}\natexlab{}.
\newblock \showarticletitle{Subjective and Objective Analysis of Streamed
  Gaming Videos}.
\newblock \bibinfo{journal}{\emph{arXiv preprint arXiv:2203.12824}}
  (\bibinfo{year}{2022}).
\newblock


\bibitem[Zadtootaghaj et~al\mbox{.}(2018)]%
        {zadtootaghaj2018nr}
\bibfield{author}{\bibinfo{person}{Saman Zadtootaghaj},
  \bibinfo{person}{Nabajeet Barman}, \bibinfo{person}{Steven Schmidt},
  \bibinfo{person}{Maria~G Martini}, {and} \bibinfo{person}{Sebastian
  M{\"o}ller}.} \bibinfo{year}{2018}\natexlab{}.
\newblock \showarticletitle{NR-GVQM: A no reference gaming video quality
  metric}. In \bibinfo{booktitle}{\emph{2018 IEEE International Symposium on
  Multimedia (ISM)}}. IEEE, \bibinfo{pages}{131--134}.
\newblock


\bibitem[Zadtootaghaj et~al\mbox{.}(2020)]%
        {zadtootaghaj2020quality}
\bibfield{author}{\bibinfo{person}{Saman Zadtootaghaj}, \bibinfo{person}{Steven
  Schmidt}, \bibinfo{person}{Saeed~Shafiee Sabet}, \bibinfo{person}{Sebastian
  M{\"o}ller}, {and} \bibinfo{person}{Carsten Griwodz}.}
  \bibinfo{year}{2020}\natexlab{}.
\newblock \showarticletitle{Quality estimation models for gaming video
  streaming services using perceptual video quality dimensions}. In
  \bibinfo{booktitle}{\emph{Proceedings of the 11th ACM Multimedia Systems
  Conference}}. \bibinfo{pages}{213--224}.
\newblock


\bibitem[Zeiler and Fergus(2014)]%
        {zeiler2014visualizing}
\bibfield{author}{\bibinfo{person}{Matthew~D Zeiler} {and} \bibinfo{person}{Rob
  Fergus}.} \bibinfo{year}{2014}\natexlab{}.
\newblock \showarticletitle{Visualizing and understanding convolutional
  networks}. In \bibinfo{booktitle}{\emph{European conference on computer
  vision}}. Springer, \bibinfo{pages}{818--833}.
\newblock


\bibitem[Zhang et~al\mbox{.}(2021a)]%
        {zhang2021continual}
\bibfield{author}{\bibinfo{person}{Weixia Zhang}, \bibinfo{person}{Dingquan
  Li}, \bibinfo{person}{Chao Ma}, \bibinfo{person}{Guangtao Zhai},
  \bibinfo{person}{Xiaokang Yang}, {and} \bibinfo{person}{Kede Ma}.}
  \bibinfo{year}{2021}\natexlab{a}.
\newblock \showarticletitle{Continual learning for blind image quality
  assessment}.
\newblock \bibinfo{journal}{\emph{arXiv preprint arXiv:2102.09717}}
  (\bibinfo{year}{2021}).
\newblock


\bibitem[Zhang et~al\mbox{.}(2021b)]%
        {zhang2021task}
\bibfield{author}{\bibinfo{person}{Weixia Zhang}, \bibinfo{person}{Kede Ma},
  \bibinfo{person}{Guangtao Zhai}, {and} \bibinfo{person}{Xiaokang Yang}.}
  \bibinfo{year}{2021}\natexlab{b}.
\newblock \showarticletitle{Task-specific normalization for continual learning
  of blind image quality models}.
\newblock \bibinfo{journal}{\emph{arXiv preprint arXiv:2107.13429}}
  (\bibinfo{year}{2021}).
\newblock


\bibitem[Zhang et~al\mbox{.}(2021c)]%
        {zhang2021uncertainty}
\bibfield{author}{\bibinfo{person}{Weixia Zhang}, \bibinfo{person}{Kede Ma},
  \bibinfo{person}{Guangtao Zhai}, {and} \bibinfo{person}{Xiaokang Yang}.}
  \bibinfo{year}{2021}\natexlab{c}.
\newblock \showarticletitle{Uncertainty-aware blind image quality assessment in
  the laboratory and wild}.
\newblock \bibinfo{journal}{\emph{IEEE Transactions on Image Processing}}
  \bibinfo{volume}{30} (\bibinfo{year}{2021}), \bibinfo{pages}{3474--3486}.
\newblock


\bibitem[Zheng et~al\mbox{.}(2022a)]%
        {zheng2022no}
\bibfield{author}{\bibinfo{person}{Qi Zheng}, \bibinfo{person}{Zhengzhong Tu},
  \bibinfo{person}{Yibo Fan}, \bibinfo{person}{Xiaoyang Zeng}, {and}
  \bibinfo{person}{Alan~C Bovik}.} \bibinfo{year}{2022}\natexlab{a}.
\newblock \showarticletitle{No-Reference Quality Assessment of Variable
  Frame-Rate Videos Using Temporal Bandpass Statistics}. In
  \bibinfo{booktitle}{\emph{ICASSP 2022-2022 IEEE International Conference on
  Acoustics, Speech and Signal Processing (ICASSP)}}. IEEE,
  \bibinfo{pages}{1795--1799}.
\newblock


\bibitem[Zheng et~al\mbox{.}(2022b)]%
        {zheng2022faver}
\bibfield{author}{\bibinfo{person}{Qi Zheng}, \bibinfo{person}{Zhengzhong Tu},
  \bibinfo{person}{Pavan~C Madhusudana}, \bibinfo{person}{Xiaoyang Zeng},
  \bibinfo{person}{Alan~C Bovik}, {and} \bibinfo{person}{Yibo Fan}.}
  \bibinfo{year}{2022}\natexlab{b}.
\newblock \showarticletitle{FAVER: Blind Quality Prediction of Variable Frame
  Rate Videos}.
\newblock \bibinfo{journal}{\emph{arXiv preprint arXiv:2201.01492}}
  (\bibinfo{year}{2022}).
\newblock


\end{thebibliography}

\end{document}